\documentclass{ceurart}

\usepackage{listings}
\usepackage{nicefrac}
\usepackage{inconsolata}
\usepackage{enumitem}

\begin{document}

\copyrightyear{2026}
\copyrightclause{Copyright for this paper by its authors.
  Use permitted under Creative Commons License Attribution 4.0
  International (CC BY 4.0).}

\conference{CLEF 2026 Working Notes, 21 -- 24 September 2026, Jena, Germany}

\title{Nika at ImageCLEF 2026 Task on Multimodal Reasoning:
More Tokens, Fewer Trees --- Test-Time Scaling for Small
VLMs on Multilingual Visual MCQ}

\title[mode=sub]{ImageCLEF Lab at CLEF 2026}

\author[1]{Spiros Baxevanakis}[%
  email=spiros.baxevanakis@student.uva.nl,
]
\fnmark[1]
\author[1]{Peng-Jian Yang}[%
  email=lesterpjy@gmail.com,
]
\fnmark[1]

\address[1]{University of Amsterdam, Science Park 904,
  1098 XH Amsterdam, The Netherlands}

\fntext[1]{These authors contributed equally.}

\begin{abstract}
Test-time scaling (TTS) reliably improves reasoning in large language models, but whether it transfers to small open vision-language models remains unclear. We examine this on EXAMS-V, a multilingual visual multiple-choice benchmark, comparing self-consistency, describe-then-reason with PRM-guided beam search, and two post-hoc selectors across Qwen2.5-VL-7B-Instruct and Qwen3.5-4B. What matters is the conditions under which TTS runs, not the search or verification machinery. The largest factor is parseability: an early prompt format left many chains reasoning correctly yet never committing to an answer letter, which a standard answer cue and a guided repair step largely remove. A larger decoding budget removes the rest: raising the per-chain token limit from 1k to 2k recovers 3.7 pp, whereas sampling more chains (8 to 16) adds only 0.15 pp. Once chains have room to finish, elaborate methods contribute little: PRM-guided beam search trails plain self-consistency by 0.39 pp at over eight times the cost, and neither a training-free generative critic nor a trained multimodal PRM beats majority vote across both policies. The largest gain comes instead from the policy model itself ($+11.4$ pp). Our best configuration reaches 84.1\% on the held-out ImageCLEF 2026 test split, ranking first on the Visual MCQ leaderboard.
\end{abstract}


\maketitle

\section{Introduction}
\label{sec:intro}

Multimodal reasoning remains a challenge for vision--language models (VLMs) whenever questions combine structured visual content with multi-step inference across languages. Real exam questions are a natural stress test of this combination, and the EXAMS-V benchmark~\citep{das2024examsv} spans thirteen languages and twenty school subjects rendered as text-in-image panels. The 2026 ImageCLEF multimodal reasoning shared task~\citep{ImageCLEF2026, ImageCLEFMultimodalReasoningTaskOverview2026} fixes the practical constraints: a single A40 GPU, open-weight policies with at most 7B parameters, and a reproducibility mandate.

The natural question is whether the inference-time techniques that advanced text-only language models~\citep{wang2023selfconsistency, lightman2024verify, snell2025compute} transfer to open VLMs under this budget. The evidence is cautionary: self-refinement degrades open-source VLMs at this scale~\citep{ahmadpour2025limits}, long sequential ``thinking'' often underperforms parallel sampling on small policies~\citep{ghosal2025mirage, gema2025inversescaling}, and discriminative reward models trained on mathematics fail to transfer to humanities and non-English content~\citep{zeng2025versaprm, Lee2025RethinkingRM}. A viable pipeline must externalise selection, prefer parallel sampling to long single-chain reasoning, and avoid iterative self-refinement loops.

We investigate whether describe-then-reason with PRM-guided search beats flat self-consistency; how accuracy scales along two axes (chain count and per-chain token budget) under a single-A40 envelope~\citep{snell2025compute}; where TTS helps and fails across subjects and languages~\citep{ahmadpour2025limits}; and whether a training-free generative critic can beat majority vote on low-agreement questions, with a cross-model PRM~\citep{ong2025qwenvlprm} controlling for sycophantic self-scoring~\citep{panickssery2024selfrecognition}. The latter extends generative-verifier findings~\citep{zhao2025genprm, zhang2025gmprm, kuang2025timprm} to a multimodal, multilingual setting under a ${\le}7$B budget not previously studied to our knowledge.

\paragraph{Summary of findings.} 
Across all four dimensions, the policy substrate and engineering details (parser format, decoding budget) dominate any search or selector method. Our best configuration, Qwen3.5-4B at SC-$N{=}16$ with a $2{,}048$-token budget, reaches $81.6\%$ on the full validation set. \footnote{Code available at: \url{https://github.com/lesterpjy/tts-small-vlm}.}

\section{Related Work}
\label{sec:related}

\paragraph{Test-time scaling: theory and limits.} \citet{snell2025compute} formalize TTS as a two-axis allocation (width: parallel chains; depth: per-chain revision or search), proving compute-optimal allocation can match naive Best-of-$N$ at less cost~\citep{liu2025computeoptimal, wu2025inferencescaling}. \citet{setlur2025verification} show the verifier advantage grows with reasoning depth, motivating an external selector; converging findings~\citep{ghosal2025mirage, gema2025inversescaling, huang2024selfcorrect, ahmadpour2025limits} argue for parallel sampling and against self-correction at our policy scale.

\paragraph{Multimodal verification and PRM generalization.} 
A \emph{process reward model} (PRM) assigns each reasoning step a probability $P(+)$ that the step is correct~\citep{lightman2024verify}; it is useful only when $P(+)$ sharply separates correct from incorrect steps. Two multimodal PRMs supersede \textsc{VisualPRM-8B}~\citep{wang2025visualprm}: \textsc{Athena-PRM-7B}~\citep{wang2025athenaprm} and \textsc{Qwen-VL-PRM-7B}~\citep{ong2025qwenvlprm}, which separates perception from reasoning errors. On the text side, discriminative math-trained PRMs fail out of distribution~\citep{zeng2025versaprm, Lee2025RethinkingRM, Chen2025FromMR} while generative verifiers transfer more reliably~\citep{zhao2025genprm, zhang2025gmprm, kuang2025timprm}. Prior EXAMS-V work~\citep{ahmed2025msa, krazheva2025contextdrift} informs our scaffold; on the search axis, \textsc{PRM-BAS}~\citep{hu2025prmbas} dominates step-level Best-of-$N$ at matched compute.

\section{Method}
\label{sec:method}

We evaluate two pipelines: describe-then-reason with PRM-guided beam search and a three-way selector contrast, and flat self-consistency with a large token budget.

\paragraph{Policy and data.}
We evaluate two open-weights policies: \texttt{Qwen/Qwen2.5-VL-7B-Instruct} (7B parameters, bfloat16) for the baseline and ablation configurations, and \texttt{Qwen/Qwen3.5-4B} (4B parameters, bfloat16), a smaller but newer-generation policy, for the headline configuration. We cap image resolution at \texttt{max\_pixels}${=}1{,}003{,}520$ (${\approx}4$k vision tokens after Qwen smart-resize), balancing legibility of text-in-image exam panels against context budget: at \texttt{max\_model\_len}${=}16{,}384$, this leaves ${\sim}12$k tokens for the prompt and generated reasoning. Inference uses vLLM~\citep{kwon2023vllm} with batched parallel decoding (\texttt{SamplingParams(n=N)}) and automatic prefix caching. All experiments run on a single A100-40\,GB GPU, which matches the shared task's A40 memory constraint. We evaluate on the full EXAMS-V validation split (4{,}651 questions, 11 languages) for headline numbers and a 200-question $(\text{language},\text{subject})$-stratified subset (seed 42, minimum ten per language) for ablations whose full-validation cost is prohibitive.

\paragraph{Describe-then-reason.}
Following~\citet{ahmed2025msa}, the describe-then-reason pipeline first samples $N$ image-grounded descriptions of the question at $T{=}0.7$. For each description, with the image removed, the policy generates $M$ reasoning chains in the flat variant, or reasoning step-by-step via PRM-BAS-style beam annealing~\citep{hu2025prmbas} in the search variant: an initial beam of width $B_0$ is expanded with $B$ continuations per surviving beam, each step is scored by Qwen-VL-PRM-7B~\citep{ong2025qwenvlprm}, which retains access to the original image, and the beam width is halved whenever the score range among the top beams exceeds a threshold $\tau$. Beams terminate on the answer cue or at a maximum depth $d$. Across all (description $\times$ terminal beam) candidates, a final selector picks the answer. In our experiments the flat variant uses $N{=}4$, $M{=}4$ ($20$ calls/q, bracketing SC-$N{=}8$ from above to test whether structured decomposition helps at higher compute); the search variant halves descriptions to $N{=}2$ because beam expansion is expensive, and adopts the hyperparameters reported in~\citet{hu2025prmbas}: $B_0{=}4$, $B{=}2$, $\tau{=}0.05$, $d{=}6$ (${\sim}13$ calls/q).

\paragraph{Selectors.}
Three selectors are applied post hoc to the same chain pool. \emph{Majority vote}: most common answer letter with log-probability tie-breaking (no extra inference). A \emph{training-free generative critic} inspired by~\citet{zhang2025gmprm} scores each chain on three axes (Appendix~\ref{app:critic-prompt}). A \emph{discriminative PRM} (Qwen-VL-PRM-7B;~\citealp{ong2025qwenvlprm}), a separately trained model, scores each chain in one-shot mode (Appendix~\ref{app:prm-mode}), controlling for sycophantic self-preference~\citep{panickssery2024selfrecognition}. Both models co-reside on the A100-40\,GB at bfloat16. Each selector uses a skip rule~\citep{wang2025ton}: high-confidence majorities (${\ge}5/8$ agreement) pass through; only the weak (${\le}4/8$) and all-unparseable tiers are rescored.

\paragraph{Self-consistency.}
The simpler pipeline samples $N \in \{8, 16\}$ chains from Qwen3.5-4B at $T{=}0.7$ with $\texttt{max\_new\_tokens}{=}2048$, conditioned on a CoT prompt ending with the MMMU closer~\citep{yue2024mmmu,yue2025mmmupro}. A regex extracts the answer letter (mapping Cyrillic labels to A--E for multilingual coverage). The final answer is the majority letter across parseable chains, with log-probability tie-breaking. We additionally sweep $T \in \{0.3, 0.5, 0.7, 0.9\}$ at full validation scale.

\paragraph{Guided parse repair.}
When every chain for a question fails to produce an answer letter ($2$--$18\%$ of questions depending on budget), each chain's reasoning is concatenated with the cue \texttt{\textbackslash n\textbackslash nAnswer:~} and a single token is decoded under vLLM's \texttt{guided\_choice} constraint over $\{A, B, C, D, E\}$. The majority across all committed letters (original plus repaired) is the final answer. This eliminates parse failures by construction.

\paragraph{Stratification and evaluation.}
We partition accuracy by subject and language; per-subject results are reported only for cells with $n{\ge}50$. Pairwise method comparisons use McNemar's paired test with Bonferroni correction for the number of reported comparisons.

\section{Experiments}
\label{sec:experiments}

We evaluate seventeen configurations across two policies, three search strategies, three selectors, and two scaling axes. Table~\ref{tab:arms} summarises them; the remaining sections report findings ordered by headline result rather than by configuration. Unless noted, all headline numbers are on the full EXAMS-V validation set ($n{=}4{,}651$); configurations marked ``dev-200'' use a 200-question stratified subset.

\begin{table*}[t]
\centering\small
\caption{Experimental configurations. ``Policy'' is Q2.5${=}$Qwen2.5-VL-7B-Instruct, Q3.5${=}$Qwen3.5-4B. Calls/q is the typical end-to-end VLM forward count. Header rows group configurations by theme. All full-validation configurations use $n{=}4{,}651$.}
\label{tab:arms}
\resizebox{\textwidth}{!}{%
\begin{tabular}{llllrl}
\toprule
\textbf{Method} & \textbf{Policy} & \textbf{Search / sampling} &
\textbf{Selector}      & \textbf{Calls/q} & \textbf{Scale} \\
\midrule
\multicolumn{6}{l}{\emph{Baselines}}\\
B0 zero-shot         & Q2.5 & guided\_choice, 1 sample      & ---                    & 1 & val-full \\
B1 chain-of-thought  & Q2.5 & 1 sample, MMMU closer         & post-hoc regex         & 1 & val-full \\
B2 self-consistency  & Q2.5 & $N{=}8$, $T{=}0.7$, 1024-tok  & majority vote          & 8 & val-full \\
B0 zero-shot         & Q3.5 & guided\_choice, 1 sample      & ---                    & 1 & val-full \\
B1 chain-of-thought  & Q3.5 & 1 sample, MMMU closer         & post-hoc regex         & 1 & val-full \\
B2 self-consistency  & Q3.5 & $N{=}8$, $T{=}0.7$, 1024-tok  & majority vote          & 8 & val-full \\
\midrule
\multicolumn{6}{l}{\emph{Search-strategy contrasts (dev-200)}}\\
S-DTR                & Q2.5 & DTR $N{=}4{,}M{=}4$           & majority vote          & 20 & dev-200 \\
S-PRM-BAS            & Q2.5 & DTR $N{=}2$ + PRM-BAS
                              ($B_0{=}4,B{=}2,d{=}6$)       & majority vote          & ${\sim}13$ & dev-200 \\
\midrule
\multicolumn{6}{l}{\emph{Scaling (Q3.5, val-full)}}\\
SC 1k-tok            & Q3.5 & $N{=}8$, $T{=}0.7$, 1024-tok  & majority vote          & 8 & val-full \\
SC 2k-tok            & Q3.5 & $N{=}8$, $T{=}0.7$, 2048-tok  & majority vote          & 8 & val-full \\
SC $N{=}16$          & Q3.5 & $N{=}16$, $T{=}0.7$, 2048-tok & majority vote          & 16 & val-full \\
Temp.\ sweep         & Q3.5 & $N{=}8$, $T{\in}\{.3,.5,.7,.9\}$, 2k & majority vote   & 8 & val-full \\
\midrule
\multicolumn{6}{l}{\emph{Selectors (full val, both pools)}}\\
R1 generative critic & Q2.5 & SC pool                       & GM-PRM-style critic    & ${+}3$/ch & val-full \\
R2 Qwen-VL-PRM       & Q2.5 & SC pool                       & PRM 1-shot + skip      & ${+}1$/ch & val-full \\
R1 generative critic & Q3.5 & SC-2k pool                    & GM-PRM-style critic    & ${+}3$/ch & val-full \\
R2 Qwen-VL-PRM       & Q3.5 & SC-2k pool                    & PRM 1-shot + skip      & ${+}1$/ch & val-full \\
\midrule
\multicolumn{6}{l}{\emph{Best configuration}}\\
S-best               & Q3.5 & $N{=}16$, $T{=}0.7$, 2048-tok & majority + guided repair & ${\le}16$ & val-full \\
\bottomrule
\end{tabular}}
\end{table*}

\section{Results and Discussion}
\label{sec:results}

We report results grouped by headline finding.

\subsection{The TTS progression and the parser artifact}
\label{sec:ladder}

\begin{table}[t]
\centering\footnotesize
\caption{TTS progression on full validation ($n{=}4{,}651$). All Q2.5 pairs are Bonferroni-significant at $\alpha{=}0.017$. Q3.5 SC rows show the effect of doubling the token budget. P.-fail: fraction of questions with all chains unparseable.}
\label{tab:progression}
\setlength{\tabcolsep}{3pt}
\begin{tabular}{llrlrr}
\toprule
\textbf{Policy} & \textbf{Method} & \textbf{Acc.} &
\textbf{95\% CI} & \textbf{P.-fail} & \textbf{C/q} \\
\midrule
Q2.5 & Zero-shot        & 56.83 & {\scriptsize[55.4, 58.2]} & 0.0\%  & 1 \\
Q2.5 & CoT              & 63.10 & {\scriptsize[61.7, 64.5]} & 2.0\%  & 1 \\
Q2.5 & SC-$N{=}8$ (1k)  & 66.42 & {\scriptsize[65.0, 67.8]} & 0.2\%  & 8 \\
\midrule
Q3.5 & Zero-shot        & 57.11 & {\scriptsize[55.7, 58.5]} & 0.0\%  & 1 \\
Q3.5 & CoT              & 69.66 & {\scriptsize[68.3, 71.0]} & 16.2\% & 1 \\
Q3.5 & SC-$N{=}8$ (1k)  & 77.81 & {\scriptsize[76.6, 79.0]} & 18.4\% & 8 \\
Q3.5 & SC-$N{=}8$ (2k)  & 81.49 & {\scriptsize[80.4, 82.6]} & 2.7\%  & 8 \\
Q3.5 & \textbf{SC-$N{=}16$ (2k)} & \textbf{81.64} & {\scriptsize\textbf{[80.5, 82.7]}} & \textbf{1.9\%} & 16 \\
\bottomrule
\end{tabular}
\end{table}

The three-stage TTS progression (zero-shot, chain-of-thought, self-consistency) holds for both policies at full validation scale (Table~\ref{tab:progression}). Each step yields a significant gain under Qwen2.5-VL-7B, and the progression is more pronounced under Qwen3.5-4B, where the full span from zero-shot to SC-$N{=}8$ (2k) exceeds 24~pp.

However, the relative contribution of each stage shifts with the policy. Under Qwen2.5, the CoT-to-SC margin is modest ($+3.3$~pp), suggesting that eight samples add limited signal once CoT prompting is in place. Under Qwen3.5, the margin widens markedly: CoT accounts for roughly half the total gain, and the remaining half comes from parallel sampling and the token-budget increase (Section~\ref{sec:scaling}).

\paragraph{A parser artifact masked part of the CoT gain.}
An early prompt format left ${\sim}9\%$ of development chains unparseable. Switching to the MMMU-standard closer~\citep{yue2024mmmu,yue2025mmmupro} cut parse failures below $2\%$ and raised dev accuracy by ${\sim}6$~pp; the fix corroborates at val scale ($+0.82$~pp on Q2.5 SC-$N{=}8$). The methodological lesson: what looks like reasoning failure can be an extraction failure, and TTS studies that do not control for parseability risk attributing engineering artifacts to the scaling method.

Under Q3.5, every step is significant: CoT over zero-shot gains $+12.6$~pp (non-overlapping CIs), the 1k$\to$2k budget increase recovers $+3.7$~pp ($171$ additional correct answers), and the $N{=}8 \to N{=}16$ step adds only $+0.15$~pp ($7$ questions, within sampling noise). Q3.5 CoT itself suffers $16.2\%$ parse-fail from the same truncation that affects SC ($18.4\%$ at 1k), so the reported $69.66\%$ underestimates single-chain accuracy and the CoT-to-SC(1k) gain conflates truncation absorption with diversity benefit.

\subsection{Structured search underperforms flat sampling}
\label{sec:dtr-bas}

Dev-200 CIs overlap between PRM-BAS, DTR, and flat SC-$N{=}8$ (Table~\ref{tab:dev200}, Appendix~\ref{app:dev200-search}). A val-scale run of PRM-BAS under Qwen3.5-4B ($N{=}2$, $B_0{=}4$, $B{=}2$, $\tau{=}0.05$, $d{=}6$; $n{=}4{,}319$, $93\%$ of validation) resolves the ranking in favour of SC: PRM-BAS reaches $80.74\%$ against SC majority at $81.13\%$ on the same questions ($-0.39$~pp). Of the $14.6\%$ of questions where PRM-BAS produces a different answer than SC, $233$ are corrections and $250$ are regressions, for a net loss of $17$ questions. Two bottlenecks interact: in the DTR pipeline the reasoning stage operates on the text description alone (no image access), so visual detail lost at the description step cannot be recovered downstream; beam search compounds this loss because the PRM scoring each step (though it retains image access) lacks sufficient discriminative signal to steer search toward the missing detail.

\paragraph{Diagnosis.}
\label{sec:perception} 
\citet{hu2025prmbas} show that PRM-guided beam-annealing search dominates flat Best-of-$N$ on multimodal math benchmarks. Our result inverts this: on multilingual visual MCQ, PRM-BAS underperforms flat SC by $-0.39$~pp at $8.7\times$ the cost. Two factors explain the discrepancy. First, the PRM's per-step P($+$) saturates ($0.962$ on descriptions, $0.849$ on reasoning), collapsing search diversity: $71.9\%$ of val questions have all surviving beams agreeing on the same letter (Shannon entropy $0.259/2.322$ bits, $89\%$ reduction from uniform); the $8.7\times$ cost yields near-zero answer diversity. Second, slicing by SC agreement tier (Appendix~\ref{app:agreement}) reveals that on high-confidence questions (${\ge}7/8$, $91.8\%$ SC accuracy), PRM-BAS regresses by $-2.16$~pp ($48$ corrections vs.\ $119$ regressions), dominating the overall deficit. Only on the $108$ all-unparseable questions does PRM-BAS produce a clear positive ($37$ correct, $34.3\%$). The P($+$) asymmetry identifies reasoning as the structural bottleneck (discussed in Section~\ref{sec:limitations}).

\subsection{Token budget dominates chain count}
\label{sec:scaling}

We sweep two scaling axes (per-chain token budget and number of sampled chains), first on the 200-question development subset, then on the full validation set (Table~\ref{tab:scaling}).

\begin{table}[t]
\centering\footnotesize
\caption{Two-axis scaling on Qwen3.5-4B, MMMU closer.
\emph{Top}: dev-200 exploration. \emph{Bottom}: val-full ($n{=}4{,}651$) validation of the key points. Dev-200's $+2$~pp $N$-scaling signal collapses at val-full.}
\label{tab:scaling}
\setlength{\tabcolsep}{3.5pt}
\begin{tabular}{llrrr}
\toprule
\textbf{Scale} & \textbf{Config.}  & \textbf{Acc.} &
\textbf{P.-fail} & \textbf{$s$/q} \\
\midrule
\multicolumn{5}{l}{\emph{Dev-200 exploration}} \\
dev & $N{=}8$, $T{=}0.7$, 1024-tok      & ${\sim}65$    & 18.0\% & ${\sim}7$  \\
dev & $N{=}8$, $T{=}0.7$, 2048-tok      & 80.0          & 2.5\%  & ${\sim}14$ \\
dev & $N{=}8$, $T{=}0.7$, 3072-tok      & 81.0          & 1.5\%  & ${\sim}21$ \\
dev & \textbf{$N{=}16$, $T{=}0.7$, 2048-tok} & \textbf{82.0} & \textbf{2.5\%} & ${\sim}$\textbf{25} \\
\midrule
\multicolumn{5}{l}{\emph{Val-full validation ($n{=}4{,}651$)}} \\
val & $N{=}8$, $T{=}0.7$, 1024-tok      & 77.81         & 18.4\% & 8.8  \\
val & $N{=}8$, $T{=}0.7$, 2048-tok      & 81.49         & 2.4\%  & 11.8 \\
val & \textbf{$N{=}16$, $T{=}0.7$, 2048-tok} & \textbf{81.64} & \textbf{1.9\%} & \textbf{14.6} \\
\midrule
\multicolumn{5}{l}{\emph{Temperature sweep (val-full, $N{=}8$, 2048-tok)}} \\
val & $T{=}0.3$                          & 80.84         & 2.7\%  & 11.2 \\
val & $T{=}0.5$                          & 81.01         & 2.6\%  & 11.5 \\
val & \textbf{$T{=}0.7$}                 & \textbf{81.49}& \textbf{2.4\%} & \textbf{11.8} \\
val & $T{=}0.9$                          & 81.04         & 3.0\%  & 12.3 \\
\bottomrule
\end{tabular}
\end{table}

\begin{figure*}[t]
\centering
\includegraphics[width=\textwidth]{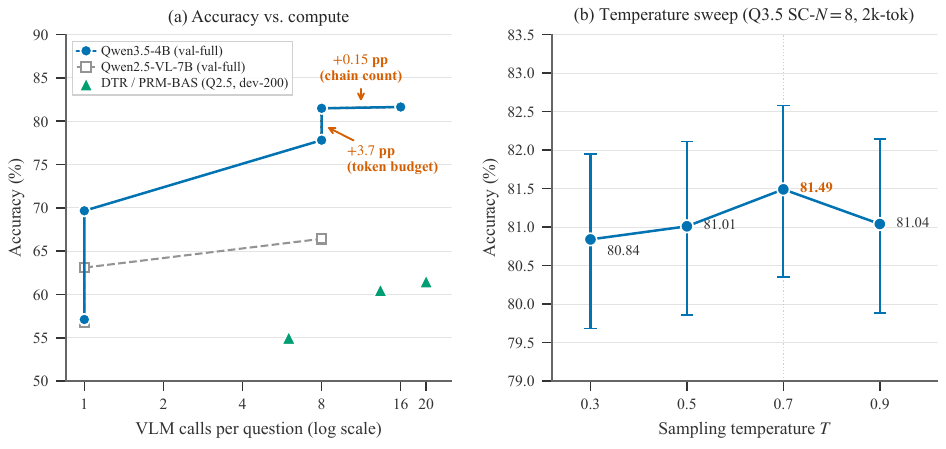}
\caption{(a)~Accuracy vs.\ VLM calls per question. At $N{=}8$, the token-budget arrow marks $+3.7$~pp from 1k$\to$2k tokens at constant call count; the chain-count arrow marks $+0.15$~pp from doubling $N$ at constant budget. DTR search points (dev-200, Q2.5) sit below flat SC at higher compute. (b)~Temperature sweep on Q3.5 SC-$N{=}8$ (2k-tok, val-full). Error bars are 95\% Wilson CIs.}
\label{fig:scaling}
\end{figure*}

\paragraph{Token budget dominates chain count.}
Doubling the budget from 1k to 2k tokens recovers $+3.7$~pp (Figure~\ref{fig:scaling}a): chains generate valid reasoning but are truncated before emitting an answer letter, producing parse failures rather than reasoning failures. Doubling chains ($N{=}8 \to 16$) adds only $+0.15$~pp. The asymmetry is structural: ${\sim}60\%$ of questions already reach strong agreement at $N{=}8$, while the weak-agreement tail (${\le}4/8$, $23\%$) has below-chance accuracy that additional sampling does not correct. (Appendix~\ref{app:agreement}). For compute-constrained TTS, ensuring chains run to completion matters more than sampling more of them.

\paragraph{Temperature sensitivity.}
An inverted-U optimum at $T{=}0.7$ replicates from dev-200 to full validation (Figure~\ref{fig:scaling}b), but the spread compresses to $0.65$~pp at scale, suggesting the policy's sampling diversity is already near optimal.

\subsection{Guided repair closes the parse-fail residue}
\label{sec:repair}
After the budget fix, val SC-$N{=}8$ still has $124$ all-unparseable questions ($2.67\%$); the test split ($n{=}1{,}117$) has $82$ ($7.34\%$), elevated because the test mix is $44.8\%$ English and $30.6\%$ Chinese. Guided repair drives both rates to zero at one decoded token per affected chain. The repair targets valid reasoning that was never committed to an answer letter, a failure mode that chain scaling cannot reach, recovering more than doubling $N$ ($+0.15$~pp val) at a fraction of the compute.

\subsection{Policy choice dominates search and selection}
\label{sec:policy}

Switching from Qwen2.5-VL-7B to Qwen3.5-4B at matched settings yields $+11.4$~pp (Table~\ref{tab:progression}, SC-$N{=}8$ rows); with the budget increase the gap widens to $15.1$~pp and all eleven languages improve by ${\ge}3$~pp (Table~\ref{tab:lang}). Qwen3.5-4B is \emph{smaller}; the gain reflects newer-generation training, consistent with~\citet{ahmadpour2025limits}. At our budget scale, upgrading the policy dominates any inference-time strategy.

\subsection{Selectors yield a null result on both pools}
\label{sec:selectors}

We test whether a post-hoc selector improves on majority vote. Two selectors are compared: a \emph{training-free generative critic} (three rubric axes; Appendix~\ref{app:critic-prompt}) and a \emph{discriminative PRM} (Qwen-VL-PRM-7B; one-shot scoring). Both apply a skip rule preserving high-confidence majorities. We evaluate on two pools of different baseline strength ($66\%$ Q2.5, $81\%$ Q3.5) to control for ceiling effects (Table~\ref{tab:selectors}).

\begin{table}[t]
\centering\footnotesize
\caption{Selectors on full validation ($n{=}4{,}651$). Both selectors are tested on two candidate pools (Q2.5 SC-$N{=}8$ and Q3.5 SC-$N{=}8$ at 2048 tokens). All use the skip-on-high rule; R2 uses one-shot PRM scoring. The Q2.5 pool predates the MMMU closer (Section~\ref{sec:ladder}); its majority anchor is therefore $65.60\%$ rather than the post-closer $66.42\%$ in Table~\ref{tab:progression}.}
\label{tab:selectors}
\setlength{\tabcolsep}{3.5pt}
\begin{tabular}{llrr}
\toprule
\textbf{Pool} & \textbf{Selector} & \textbf{Acc.} & \textbf{$\Delta$} \\
\midrule
Q2.5 & SC majority (anchor)     & 65.60\% & ---            \\
Q2.5 & R1 generative critic     & 65.60\% & $+0.00$ pp     \\
Q2.5 & R2 PRM (1-shot)          & 66.05\% & $+0.45$ pp$^{\text{n.s.}}$ \\
\midrule
Q3.5 & SC majority (anchor)     & 81.49\% & ---            \\
Q3.5 & R2 PRM (1-shot)          & 81.40\% & $-0.09$ pp     \\
Q3.5 & R1 generative critic     & 81.01\% & $-0.47$ pp     \\
\bottomrule
\end{tabular}
\\[2pt]
{\scriptsize $^{\text{n.s.}}$Q2.5 R2: McNemar $\chi^2{=}1.80$,
$p{\approx}0.18$. Q3.5: both negative, not tested.}
\end{table}

\paragraph{Cross-pool replication.}
The null replicates across both policies: on Q2.5 ($66\%$) the critic is an exact null and the PRM gains $+0.45$~pp (n.s.); on Q3.5 ($81\%$) both turn negative (PRM: $51$ corrected vs.\ $55$ regressed; critic: $37$ vs.\ $59$). As pool accuracy grows, the majority-wrong tail shrinks and a near-balanced selector's net effect turns negative (Appendix~\ref{app:critic}).

\paragraph{Why both selectors fail.}
The critic exhibits self-recognition bias~\citep{panickssery2024selfrecognition}: $40\%$ of chains receive the modal score $0.75$, and the correct/incorrect gap is negligible ($0.713$ vs.\ $0.687$; Appendix~\ref{app:critic}). The PRM improves the correction rate ($37.6\%$ vs.\ $31.8\%$), partially mitigating self-preference, but cannot distinguish right from wrong overrides: the score gap between its top-ranked chain and runner-up is $0.049$ on corrections and $0.048$ on degradations, yielding a net effect indistinguishable from noise.

\subsection{Stratified analysis}
\label{sec:stratified}

\begin{figure*}[t]
\centering
\includegraphics[width=\textwidth]{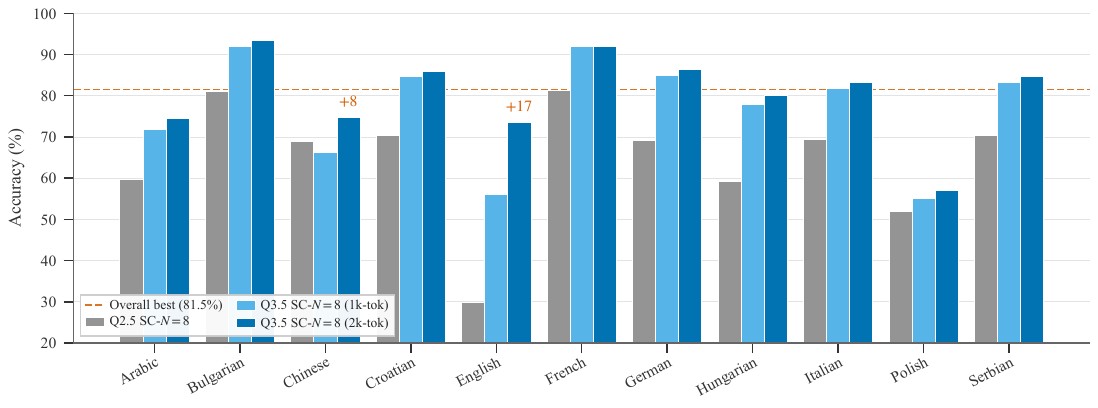}
\caption{Per-language accuracy on full validation. Red annotations mark the accuracy gain from the budget increase (1k$\to$2k) for languages where it exceeds $+5$~pp. The dashed line is the overall best (81.5\%).}
\label{fig:langs}
\end{figure*}

Figure~\ref{fig:langs} and Table~\ref{tab:lang} (Appendix~\ref{app:lang-table}) report per-language accuracy under both policies, with the Q3.5 column split by token budget to isolate the effect of the budget increase.

\paragraph{The budget increase concentrates gains on
English and Chinese.}
English gains $+17.3$~pp and Chinese $+8.5$~pp when the budget doubles; under Q2.5, over a third of English questions had all eight chains unparseable. Most other languages gain only $1$--$2$~pp and French ($92\%$) is unchanged, confirming a targeted rather than general effect.

\paragraph{Persistent below-average strata.}
Chinese, Polish, English, and Arabic remain below $81.5\%$ even after the budget increase. Under Q2.5, SC \emph{regresses} on Chinese by $2.8$~pp: a chain-agreement audit (Appendix~\ref{app:lang-agreement}) shows $22.3\%$ of Chinese questions reach unanimous $8/8$ agreement (vs.\ ${\le}15\%$ elsewhere). SC gains power when chains err independently; when errors are highly correlated, the vote amplifies the shared mistake. Languages benefiting most from SC (Bulgarian, Croatian, Hungarian, Serbian) sit at mid-accuracy where the independence condition holds.

\paragraph{Per-subject patterns.}
Per-subject accuracy (Appendix~\ref{app:subjects}) reveals a split: STEM subjects benefit most from TTS (Physics $+14$~pp, Mathematics $+20$~pp over zero-shot), while the weakest Q3.5 subjects (Social $43\%$, Islamic Studies $57.8\%$) are text-heavy. Using subject as a proxy for visual complexity (EXAMS-V lacks content-type annotations), the pattern suggests two regimes: in STEM, reasoning errors vary across chains and majority vote aggregates them away; in the weakest subjects, the bottleneck is missing domain knowledge that no amount of sampling can compensate.

\subsection{Held-out test performance}
\label{sec:test}

Our best configuration (Qwen3.5-4B at SC-$N{=}16$, $2{,}048$-token budget, with guided repair) ranks first on the official Visual MCQ leaderboard at $84.06\%$ overall, and is first on each of the six test languages (Table~\ref{tab:leaderboard}, Appendix~\ref{app:leaderboard}).

The test split ($n{=}1{,}117$) is $44.8\%$ English and $30.6\%$ Chinese, and both exceed their validation accuracy: English rises to $85.6\%$ (from $73.5\%$ val, $+12.1$~pp) and Chinese to $78.1\%$ (from $74.7\%$, $+3.4$~pp). These are the two languages where the token-budget and guided-repair fixes (Sections~\ref{sec:scaling},~\ref{sec:repair}) had the most headroom on validation, so the test result tracks with the validation diagnosis.

\section{Conclusion}
\label{sec:conclusion}

We studied test-time scaling for open VLMs on multilingual visual MCQ, comparing structured search against flat self-consistency across two policies on the full EXAMS-V validation set. TTS improves accuracy: the full progression from zero-shot to self-consistency spans over 24~pp on Qwen3.5-4B, but the gains concentrate on the per-chain token budget rather than on structured search, larger chain counts, or trained selectors: neither a generative critic nor a trained PRM beats majority vote across two policy models and two pool strengths. Stratified analysis confirms that TTS gains concentrate on mid-accuracy languages where the policy is miscalibrated rather than weak, and that persistent below-average strata (Polish, Islamic Studies) resist SC because chains converge on the same wrong answer, leaving no diversity for the majority vote to correct.

The overarching lesson is that at our budget scale, the engineering substrate (parser format, decoding budget) and the policy choice dominate any search or selector method. Our best configuration, Qwen3.5-4B at SC-$N{=}16$ with a $2{,}048$-token budget and guided repair (Section~\ref{sec:repair}), reaches $81.6\%$ on full validation and $84.1\%$ on the held-out ImageCLEF 2026 test split, ranking first on the Visual MCQ leaderboard (Section~\ref{sec:test}). Future work should investigate whether decoupling perception from reasoning via a dedicated reasoning model can address the P($+$) asymmetry identified in Section~\ref{sec:perception}, and whether better-calibrated PRMs trained on multilingual non-mathematical content would break the selector null.

\section{Limitations}
\label{sec:limitations}
Our findings rest on two policy models, Qwen2.5-VL-7B-Instruct and Qwen3.5-4B, both from the same family, so we cannot separate behavior typical of small open VLMs from behavior specific to Qwen models. The parseability effect in particular depends on how a model emits answer letters under a given prompt format, which may differ for other open source VLM families. The 11.4 pp gain from switching policy supports our claim that the policy substrate dominates inference-time strategy, but also shows that a third model could shift the picture; whether the pattern holds across families and across the sub-7B range is untested. Ablations whose full-validation cost was prohibitive used the dev-200 subset, whose $\pm$7 pp Wilson CIs preclude ranking claims, so all headline findings rest on val-full re-runs (n=4,651). The segmented PRM null appears structural, since P($+$) ${\approx}0.85$ on reasoning steps leaves little signal once aggregated, though better per-step calibration on multilingual mixed-subject content might change this. Following the 2025 ImageCLEF winners~\citep{ahmed2025msa}, we did not fine-tune, use few-shot exemplars, or apply test-time image augmentation, all of which remain open.

\begin{acknowledgments}
Experiments were conducted on the Snellius national supercomputer, provided by the University of Amsterdam.
\end{acknowledgments}

\section*{Declaration on Generative AI}

During the preparation of this work, the authors use AI for proofreading, polishing and rephrasing sentences and paragraphs in the manuscript.

\bibliography{bibfile}

\appendix

\section{Chain-Agreement Analysis}
\label{app:agreement}

Table~\ref{tab:agreement} reports per-tier accuracy on the Q3.5 SC-$N{=}8$ (2k-tok) pool. The agreement tier is the number of chains (out of 8) that commit the same answer letter.

\begin{table}[h]
\centering\footnotesize
\caption{Chain-agreement tiers on Q3.5 SC-$N{=}8$ (2k-tok, $n{=}4{,}651$). Weak-agreement questions (${\le}4/8$) have accuracy below chance for 4-option MCQ.}
\label{tab:agreement}
\begin{tabular}{lrrr}
\toprule
\textbf{Tier} & \textbf{$n$} & \textbf{\% of val} &
\textbf{Accuracy} \\
\midrule
Unanimous ($8/8$) & 1437 & 30.9\% & 90.2\% \\
Strong ($6$--$7/8$) & 1364 & 29.3\% & 70.9\% \\
Moderate ($5/8$) & 637 & 13.7\% & 53.5\% \\
Weak (${\le}4/8$) & 1089 & 23.4\% & 41.7\% \\
All unparseable & 124 & 2.7\% & --- \\
\bottomrule
\end{tabular}
\end{table}

The skip rule used by both selectors (Section~\ref{sec:selectors}) passes through every question with at least $5/8$ agreement, i.e.\ the unanimous, strong, and moderate tiers ($73.9\%$ of validation). Only the weak and all-unparseable tiers ($26.1\%$) are rescored. On this scored subset the base accuracy is ${\sim}37\%$, leaving limited room for any selector to improve on majority vote.

\section{Generative Critic Score Distribution}
\label{app:critic}

The training-free generative critic scores each chain on three axes (step intent, visual alignment, logical soundness), each rated $1$--$5$, mapped to $[0,1]$ via $(s{-}1)/4$, and mean-aggregated. Table~\ref{tab:critic-dist} shows the score distribution across 5{,}710 scored chains from the Q2.5 SC-$N{=}8$ pool.

\begin{table}[h]
\centering\footnotesize
\caption{Generative critic score distribution. The modal score $0.75$ ($4/4/4$ axis pattern) accounts for $40\%$ of chains. The calibration slope is nearly flat: a perfect $1.0$ is correct only $40\%$ of the time.}
\label{tab:critic-dist}
\begin{tabular}{rrr}
\toprule
\textbf{Score} & \textbf{\% of chains} & \textbf{Acc.\ at score} \\
\midrule
0.00 &  0.1\% & 14.3\% \\
0.25 &  5.7\% & 27.4\% \\
0.50 &  5.0\% & 29.9\% \\
\textbf{0.75 (mode)} & \textbf{39.7\%} & \textbf{33.6\%} \\
0.83 &  8.8\% & 30.5\% \\
0.92 &  5.3\% & 33.0\% \\
1.00 &  9.8\% & 39.7\% \\
\bottomrule
\end{tabular}
\end{table}

The correct-vs-incorrect chain score gap is negligible: correct chains average $0.713$ vs.\ $0.687$ for incorrect (gap $0.026$), with an identical median of $0.75$. In $59\%$ of questions, multiple chains are tied at the top score, forcing arbitrary tie-breaking. The pattern is consistent with the self-recognition bias reported by~\citet{panickssery2024selfrecognition}: the policy rates its own outputs uniformly high regardless of correctness.

\section{Per-Language Chain Agreement}
\label{app:lang-agreement}

Table~\ref{tab:lang-agree} breaks down chain agreement by language for the Q2.5 SC-$N{=}8$ pool, revealing the structural differences that drive per-language selector performance.

\begin{table}[h]
\centering\footnotesize
\caption{Per-language chain agreement on Q2.5 SC-$N{=}8$ ($n{=}4{,}651$). English has $34\%$ all-unparseable questions; Arabic and Hungarian have the largest weak-agreement tails.}
\label{tab:lang-agree}
\setlength{\tabcolsep}{2.5pt}
\begin{tabular}{lrrrrr}
\toprule
\textbf{Lang} & $n$ & \textbf{Unan.} & \textbf{Strong} &
\textbf{Weak} & \textbf{Unparse.} \\
\midrule
Arabic    & 517 & 19.0\% & 29.2\% & 36.4\% & 0.4\% \\
Bulgarian & 400 & 40.8\% & 34.5\% & 13.2\% & 0.0\% \\
Chinese   & 600 & 22.3\% & 29.3\% & 30.2\% & 0.2\% \\
Croatian  & 585 & 39.1\% & 29.1\% & 18.5\% & 0.0\% \\
English   & 347 & 14.1\% & 10.1\% & 36.0\% & \textbf{34.0\%} \\
French    & 224 & 43.3\% & 35.3\% & 11.6\% & 0.0\% \\
German    & 279 & 30.5\% & 39.4\% & 15.8\% & 0.4\% \\
Hungarian & 535 & 20.4\% & 32.0\% & 30.1\% & 0.4\% \\
Italian   & 562 & 45.0\% & 28.5\% & 15.3\% & 0.0\% \\
Polish    & 100 & 41.0\% & 25.0\% & 20.0\% & 0.0\% \\
Serbian   & 502 & 35.7\% & 29.7\% & 19.3\% & 0.0\% \\
\bottomrule
\end{tabular}
\end{table}

English performs worst under Q2.5: only $14\%$ unanimous, $34\%$ all-unparseable, $36\%$ weak agreement; $70\%$ of English questions fall in the worst two tiers, explaining why English is the only language where SC \emph{regresses} against zero-shot under Q2.5. Arabic and Hungarian have the largest weak-agreement tails ($36\%$ and $30\%$), making them the languages where selectors have the most questions to rescore and the least signal to work with. Bulgarian, French, and Italian show consistent agreement distributions ($39$--$45\%$ unanimous, small weak tails, $80$+ \% SC accuracy).

\section{Per-Subject Accuracy}
\label{app:subjects}

Table~\ref{tab:subj} reports accuracy for the five weakest and five strongest subjects under the best Q3.5 and Q2.5 configurations respectively.

\begin{table}[h]
\centering\footnotesize
\caption{Per-subject accuracy for selected subjects ($n{\ge}50$). Bottom 5 and top 5 under Q3.5 SC-$N{=}8$ (2k-tok). Q2.5 columns: zero-shot (ZS), CoT, SC-$N{=}8$.}
\label{tab:subj}
\setlength{\tabcolsep}{2.5pt}
\begin{tabular}{lrrrr|r}
\toprule
& & \multicolumn{3}{c|}{\textbf{Q2.5}} &
\textbf{Q3.5} \\
\cmidrule(lr){3-5}\cmidrule(lr){6-6}
\textbf{Subject} & $n$ & ZS & CoT & SC & \textbf{SC 2k} \\
\midrule
\multicolumn{6}{l}{\emph{Bottom 5 (Q3.5)}} \\
Social          & 100 & 28.0 & 39.0 & 49.0 & \textbf{43.0} \\
Professional    & 100 & 47.0 & 51.0 & 52.0 & \textbf{57.0} \\
Islamic Studies &  83 & 21.7 & 45.8$^*$ & 33.7$^\downarrow$ & \textbf{57.8} \\
Agriculture     & 100 & 47.0 & 48.0 & 42.0$^\downarrow$ & \textbf{59.0} \\
Tourism         &  76 & 55.3 & 63.2 & 61.8 & \textbf{67.1} \\
\midrule
\multicolumn{6}{l}{\emph{Top 5 (Q2.5 SC, for contrast)}} \\
Psychology      &  81 & 80.2 & 91.4 & 85.2 & --- \\
Politics        & 135 & 76.3 & 76.3 & 80.0 & --- \\
Sociology       & 190 & 66.8 & 72.1 & 77.4 & --- \\
Biology         & 594 & 65.3 & 70.2 & 72.6 & --- \\
Mathematics     & 100 & 49.0 & 60.0 & 69.0 & --- \\
\bottomrule
\end{tabular}
\\[2pt]
{\scriptsize $^*$CoT $>$ SC under Q2.5 (regression). $^\downarrow$SC regresses vs.\ ZS. ``--'' $=$ Q3.5 per-subject data unavailable for non-bottom-5 subjects at time of writing.}
\end{table}

\section{PRM Scoring Mode (One-shot vs Per-step)}
\label{app:prm-mode}

On a DTR $N{=}4{,}M{=}4$ pool under Qwen2.5-VL-7B (dev-200), per-step (segmented) PRM scoring is an exact null ($63.0\% \to 63.0\%$) while one-shot scoring shows a directional $+3.0$~pp gain ($63.0\% \to 66.0\%$) and is $9.4\times$ faster (Table~\ref{tab:critic}). However, the $\pm 7$~pp Wilson CIs at $n{=}200$ overlap fully, so the difference is not significant. One-shot evaluation gives the PRM maximal context to judge chain quality holistically, avoiding the per-step error compounding diagnosed in Section~\ref{sec:dtr-bas}.

\begin{table}[h]
\centering\footnotesize
\caption{PRM scoring mode on a DTR Q2.5 pool (dev-200). CIs overlap; the one-shot gain is directional only.}
\label{tab:critic}
\setlength{\tabcolsep}{3.5pt}
\begin{tabular}{lrrrr}
\toprule
\textbf{Selector}            & \textbf{Acc.} & \textbf{95\% CI} &
\textbf{$\Delta$} & \textbf{$s$/q} \\
\midrule
DTR majority (anchor)        & 63.0\% & {\scriptsize[56.1, 69.4]} & ---       & ---  \\
PRM segmented (per-step)     & 63.0\% & {\scriptsize[56.1, 69.4]} & $+0.0$ pp & 56.3 \\
PRM flat (one-shot)          & 66.0\% & {\scriptsize[59.2, 72.2]} & $+3.0$ pp & 6.0  \\
\bottomrule
\end{tabular}
\end{table}

\section{Generative Critic Prompt}
\label{app:critic-prompt}

The critic scores each chain independently on three axes, one inference call per axis. The prompt template is:

\begin{quote}\small\ttfamily
You are a rigorous evaluator reviewing a reasoning chain written by a student answering a multiple-choice exam question. The question and its answer options are shown in the image.

Rate the chain on ONE axis: \{axis\_name\}.

Axis definition: \{axis\_definition\}

Scale: 1 (severely deficient), 2 (weak), 3 (adequate),
4 (strong), 5 (excellent).

Reasoning chain to evaluate:\\
---\\
\{chain\_text\}\\
---

Output ONLY a JSON object on a single line with exactly two fields:\\
\{"score": <integer 1-5>, "reason": "<brief one-sentence justification>"\}\\
Do not output anything else.
\end{quote}

\noindent The three axes are:

\begin{description}[style=unboxed,leftmargin=0pt,itemsep=2pt]
\item[Step intent.] Does each reasoning step address the question directly and build toward a commitment to one of the answer options? Penalise irrelevant tangents, repetition, self-doubt loops, or failure to commit to a letter.
\item[Visual alignment.] Does the chain's reasoning match what is actually visible in the image? Penalise hallucinated elements, misread values, missing visual evidence, or contradictions with the options rendered in the image.
\item[Logical soundness.] Does the final answer follow from the premises via valid logical, mathematical, or scientific steps? Penalise unjustified leaps, false equivalences, arithmetic errors, or a mismatch between the concluded letter and the reasoning that preceded it.
\end{description}

\noindent Per-axis scores are mapped to $[0,1]$ via $(s{-}1)/4$ and mean-aggregated to a single chain score.

\section{Per-Language Accuracy}
\label{app:lang-table}

\begin{table}[h]
\centering\footnotesize
\caption{Per-language accuracy on full validation ($n{=}4{,}651$). Q2.5 columns: ZS, CoT, SC-$N{=}8$ (MMMU closer). Q3.5 columns: SC-$N{=}8$ at 1024 and 2048 tokens. $\Delta_{\text{1k}\to\text{2k}}$ is the gain from the budget increase. Down-arrow marks a regression.}
\label{tab:lang}
\setlength{\tabcolsep}{2.5pt}
\begin{tabular}{lr|rrr|rrr}
\toprule
& & \multicolumn{3}{c|}{\textbf{Q2.5}} &
\multicolumn{3}{c}{\textbf{Q3.5 SC}} \\
\cmidrule(lr){3-5}\cmidrule(lr){6-8}
\textbf{Lang} & $n$ & ZS & CoT & SC & 1k & \textbf{2k} &
$\Delta_{\text{1k}\to\text{2k}}$ \\
\midrule
Arabic    & 517 & 40.8 & 56.7 & 59.8 & 71.8 & \textbf{74.5} & $+2.7$ \\
Bulgarian & 400 & 58.8 & 76.8 & 81.0 & 92.0 & \textbf{93.5} & $+1.5$ \\
Chinese   & 600 & 71.8 & 60.8$^\downarrow$ & 69.0 & 66.2 & \textbf{74.7} & $+8.5$ \\
Croatian  & 585 & 60.7 & 67.0 & 70.4 & 84.6 & \textbf{86.0} & $+1.4$ \\
English   & 347 & 36.3 & 38.0 & 30.0$^\downarrow$ & 56.2 & \textbf{73.5} & $+17.3$ \\
French    & 224 & 69.6 & 78.1 & 81.3 & 92.0 & \textbf{92.0} & $+0.0$ \\
German    & 279 & 61.7 & 68.1 & 69.2 & 84.9 & \textbf{86.4} & $+1.5$ \\
Hungarian & 535 & 49.4 & 57.8 & 59.3 & 77.9 & \textbf{80.0} & $+2.1$ \\
Italian   & 562 & 64.4 & 69.0 & 69.4 & 81.9 & \textbf{83.3} & $+1.4$ \\
Polish    & 100 & 47.0 & 51.0 & 52.0 & 55.0 & \textbf{57.0} & $+2.0$ \\
Serbian   & 502 & 56.6 & 66.3 & 70.5 & 83.3 & \textbf{84.7} & $+1.4$ \\
\bottomrule
\end{tabular}
\end{table}

\section{Dev-200 Search Comparison}
\label{app:dev200-search}

\begin{table}[h]
\centering\footnotesize
\caption{Search and verification on dev-200 under Q2.5. All rows use the same pre-MMMU closer for a matched comparison; the MMMU closer was adopted later and applied at val scale (Table~\ref{tab:progression}). At $n{=}200$ the $\pm 7$~pp Wilson intervals overlap, so dev-200 alone cannot rank the more expensive methods against SC.}
\label{tab:dev200}
\setlength{\tabcolsep}{3pt}
\begin{tabular}{lrlrr}
\toprule
\textbf{Method}                    & \textbf{Acc.} &
\textbf{95\% CI} & \textbf{Calls/q} & \textbf{$s$/q} \\
\midrule
\textbf{SC-$N{=}8$}                & \textbf{65.5\%} & {\scriptsize\textbf{[58.6, 71.8]}} & \textbf{8}    & \textbf{7.4}  \\
DTR $N{=}2{,}M{=}2$                & 55.0\% & {\scriptsize[48.1, 61.7]} & 6    & 14.1 \\
DTR $N{=}4{,}M{=}4$                & 61.5\% & {\scriptsize[54.6, 68.0]} & 20   & 28.5 \\
DTR-$N{=}2$ + PRM-BAS              & 60.5\% & {\scriptsize[53.6, 67.0]} & 13.4 & 42.3 \\
\bottomrule
\end{tabular}
\end{table}

\section{Official Leaderboard}
\label{app:leaderboard}

\begin{table*}[h]
\centering\footnotesize
\caption{Official ImageCLEF 2026 Visual MCQ leaderboard (test split, $n{=}1{,}117$): overall accuracy and per-language breakdown across the six test languages. Our submission (leaderboard handle \texttt{spirosbax}) ranks first overall and on every language.}
\label{tab:leaderboard}
\setlength{\tabcolsep}{4pt}
\begin{tabular}{rlrrrrrrr}
\toprule
\textbf{\#} & \textbf{Participant} & \textbf{Overall} &
\textbf{EN} & \textbf{BG} & \textbf{ZH} & \textbf{HR} &
\textbf{IT} & \textbf{SR} \\
\midrule
\textbf{1} & \textbf{spirosbax (ours)} & \textbf{84.06} & \textbf{85.60} & \textbf{89.09} & \textbf{78.07} & \textbf{87.72} & \textbf{90.74} & \textbf{87.04} \\
2 & DS@GT           & 79.86 & 85.20 & 86.36 & 67.25 & 85.96 & 87.04 & 83.33 \\
3 & FAU             & 71.08 & 74.80 & 69.09 & 63.45 & 73.68 & 81.48 & 75.93 \\
4 & zhaijinghe      & 61.50 & 67.60 & 61.82 & 53.51 & 61.40 & 59.26 & 57.41 \\
5 & sjaini          & 59.27 & 69.20 & 50.91 & 44.15 & 59.65 & 72.22 & 66.67 \\
6 & begyed          & 57.74 & 60.00 & 66.36 & 47.08 & 66.67 & 70.37 & 64.81 \\
7 & linxiaocan      & 55.60 & 60.40 & 54.55 & 51.17 & 50.88 & 59.26 & 42.59 \\
8 & pratikpriyanshu & 50.76 & 57.20 & 43.64 & 41.81 & 56.14 & 55.56 & 51.85 \\
9 & wether          & 46.73 & 54.60 & 56.36 & 31.87 & 45.61 & 48.15 & 48.15 \\
\bottomrule
\end{tabular}
\end{table*}

\end{document}